\documentclass[letterpaper]{article} 
\usepackage{aaai24}  
\usepackage{times}  
\usepackage{helvet}  
\usepackage{courier}  
\usepackage[hyphens]{url}  
\usepackage{graphicx} 
\usepackage{natbib}  
\usepackage{caption} 
\usepackage{algorithm}
\usepackage{algorithmic}

\usepackage{multirow}
\usepackage{amsfonts,amssymb}
\usepackage{amsmath}
\usepackage{amsthm}
\usepackage{stmaryrd}
\usepackage{graphicx}
\usepackage{float}
\usepackage{color}
\usepackage{booktabs}

\usepackage{newfloat}
\usepackage{listings}
\DeclareCaptionStyle{ruled}{labelfont=normalfont,labelsep=colon,strut=off} 
\floatstyle{ruled}
\newfloat{listing}{tb}{lst}{}
\floatname{listing}{Listing}
\title{A Positive-Unlabeled Metric Learning Framework for \\Document-Level Relation Extraction with Incomplete Labeling}
\author{
    Ye Wang\textsuperscript{\rm 1},
    Huazheng Pan\textsuperscript{\rm 1},
    Tao Zhang\textsuperscript{\rm 2},
    Wen Wu\textsuperscript{\rm 1},
    Wenxin Hu\textsuperscript{\rm 1}\thanks{Corresponding author.}
}
\affiliations{
    \textsuperscript{\rm 1}East China Normal University, Shanghai, China\\
    \textsuperscript{\rm 2}Tsinghua University, Beijing, China

    \{yewang, hzpan\}@stu.ecnu.edu.cn,
    tao-zhan20@mails.tsinghua.edu.cn\\
    wwu@cs.ecnu.edu.cn,
    wxhu@cc.ecnu.edu.cn
}

\usepackage{bibentry}

\begin{document}

\maketitle

\begin{abstract}
The goal of document-level relation extraction (RE) is to identify relations between entities that span multiple sentences. Recently, incomplete labeling in document-level RE has received increasing attention, and some studies have used methods such as positive-unlabeled learning to tackle this issue, but there is still a lot of room for improvement. Motivated by this, we propose a \underline{p}ositive-augmentation and \underline{p}ositive-mixup \underline{p}ositive-unlabeled \underline{m}etric learning framework (P$^{3}$M). Specifically, we formulate document-level RE as a metric learning problem. We aim to pull the distance closer between entity pair embedding and their corresponding relation embedding, while pushing it farther away from the none-class relation embedding. Additionally, we adapt the positive-unlabeled learning to this loss objective. In order to improve the generalizability of the model, we use dropout to augment positive samples and propose a positive-none-class mixup method. Extensive experiments show that P$^{3}$M improves the F1 score by approximately 4-10 points in document-level RE with incomplete labeling, and achieves state-of-the-art results in fully labeled scenarios. Furthermore, P$^{3}$M has also demonstrated robustness to prior estimation bias in incomplete labeled scenarios.
\end{abstract}

\section{Introduction}

Relation extraction (RE) involves identifying the relations between two entities in a given text, which is a fundamental task in information extraction. In the past, most RE research focused on extracting relations within a single sentence ~\cite{DBLP:conf/acl/MiwaB16, DBLP:conf/emnlp/Zhang0M18}. However, more recent work has begun to examine document-level RE, which involves identifying relations between entities across multiple sentences in a document ~\cite{DBLP:conf/acl/YaoYLHLLLHZS19, DBLP:conf/aaai/Zhou0M021, DBLP:conf/naacl/XuCMZ22,DBLP:conf/naacl/YuYT22,DBLP:conf/ijcai/ZhouL22}.

Previously, document-level RE focused on fully supervised scenarios. However, due to the fact that the number of entity pairs is related to the number of entities in a quadratic way, it is very difficult to fully annotate all the relations in a document. This has made the problem of incomplete labeling a common problem in document-level RE and has attracted increasing attention from researchers. ~\cite{DBLP:conf/acl/HuangH0ZF022} noticed that the popular document-level RE dataset DocRED ~\cite{DBLP:conf/acl/YaoYLHLLLHZS19} annotated using the \emph{recommend-revise} scheme contains a large number of unlabeled positive relations, i.e. false negatives. ~\cite{DBLP:conf/emnlp/Tan0BNA22} obtained a high-quality Re-DocRED dataset by supplementing the large number of missing relations in DocRED. ~\cite{DBLP:conf/emnlp/WangLHZ22} was the first to use positive-unlabeled (PU) learning, a method of learning risk estimators from positive and unlabeled data, to solve the document-level RE task with incomplete labeling and provided a powerful baseline. Despite this, they still suffer greatly from distribution bias caused by incomplete annotation of positive samples, and lack of generalization in the model.

Inspired by these studies, we propose a \underline{p}ositive-augmentation and \underline{p}ositive-mixup \underline{p}ositive-unlabeled \underline{m}etric learning framework (P$^{3}$M). Firstly, for metric learning in document-level RE, we initialize an embedding for each relation and an embedding for the none-class relation. During training, we pull the entity pair embedding closer to the corresponding relation embedding and push it away from the none-class relation embedding. We adapt this goal to the positive-unlabeled learning paradigm. Then, due to the fact that the labeled positive samples are a subset of the overall positive samples, the distribution of the labeled positive samples cannot approximate the true positive sample distribution, especially in extreme cases of incomplete labeling. To alleviate this problem, inspired by ~\cite{DBLP:conf/emnlp/GaoYC21}, we use the dropout noise ~\cite{DBLP:journals/jmlr/SrivastavaHKSS14} inherent in the model to augment the positive samples and experimentally verify the effectiveness of this augmentation.

Finally, to further enhance the model's generalization, we use mixup to interpolate between the embeddings of positive and negative samples. Under the positive-unlabeled setting, it is not possible to obtain true negative samples, i.e. unlabeled entity pairs may still have some relations. Directly interpolating between the two would introduce noise. Thanks to the metric learning framework, which puts the embeddings of none-class relation and none-class entity pairs in the same feature space, we can use the embedding of none-class relation as pseudo-negative entity pair embedding.

\begin{figure*}[!ht]
\centering
\includegraphics[width=1.0\textwidth]{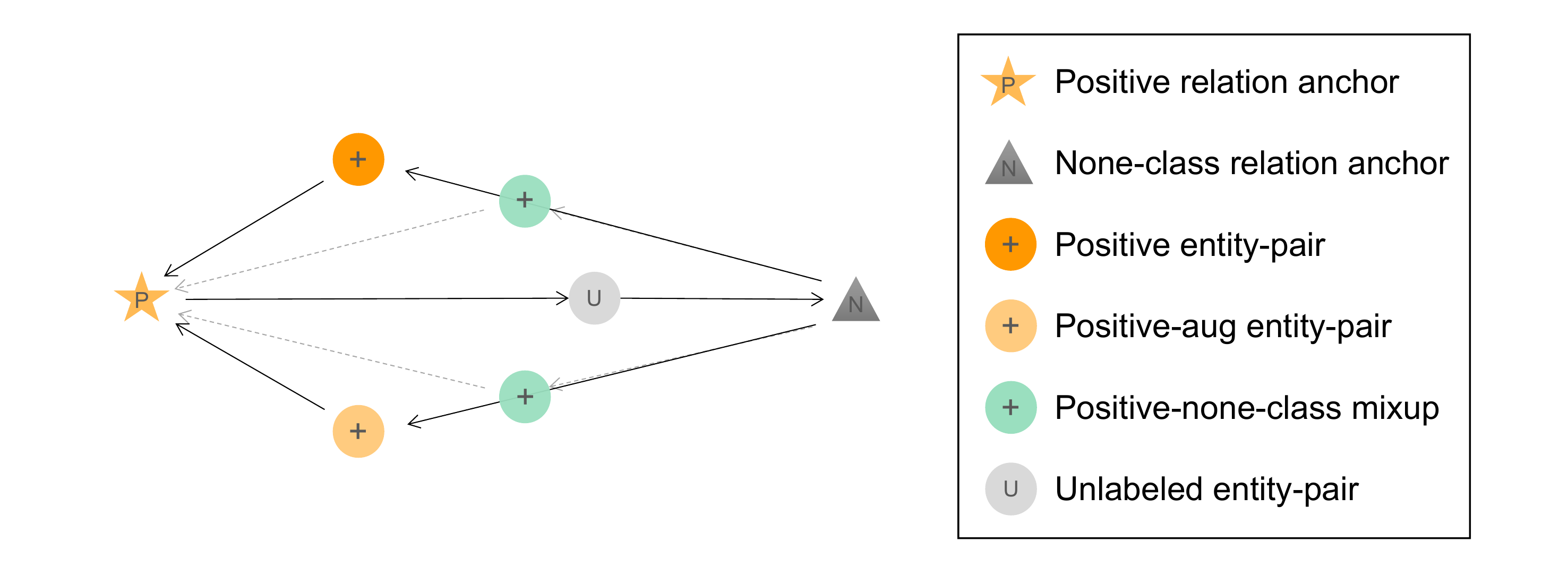} 
\caption{In the dense representation space for a specific positive relation, the P$^{3}$M framework brings the positive sample (orange circle) and its augmented embedding (light orange circle) closer to the positive relation embedding (yellow pentagram), while distancing them from the none-class relation embedding (grey triangle). The unlabeled sample (light grey circle) is distanced from the positive relation and brought closer to the none-class relation. To address scarcity of positive samples, extra positive sample embeddings (light green circles) are obtained using mixup, partially aligning them with the positive relation and distancing them from the none-class relation.}
\label{fig1}
\end{figure*}

We conduct experiments on the DocRED ~\cite{DBLP:conf/acl/YaoYLHLLLHZS19} dataset under incomplete labeling and extreme incomplete labeling settings, as well as the ChemDisGene ~\cite{DBLP:conf/lrec/ZhangMTM22} dataset in the biomedical domain. We improve the F1 score by about 4-10 points compared to the baseline, demonstrating the effectiveness of our proposed P$^{3}$M method. We also conduct experiments in the fully labeled scenario and achieved the best results. Finally, experiments under different estimated priors demonstrate the robustness of our method to prior estimation bias, which is greatly beneficial for the application of P$^{3}$M in real-world scenarios. The contributions of this paper can be summarized as follows:\footnote{Code is available at \url{https://github.com/www-Ye/P3M}}

\begin{itemize}
\item We propose a positive-unlabeled metric learning framework that adapts the metric learning objective to the positive-unlabeled learning paradigm in document-level RE.
\item We use the dropout noise inherent in the model to augment the positive samples, expanding the distribution of the positive samples.
\item We use mixup to interpolate between the embeddings of positive entity pairs and none-class relation, further enhancing the model's generalization.
\item Experiments show that our method achieves the state-of-the-art results in various incomplete labeling settings and in fully labeled scenario, as well as robustness to prior estimation bias.
\end{itemize}

\section{Methodology}
In this section, we introduce the details of P$^{3}$M. Firstly, we propose positive-unlabeled metric learning for document-level RE. Then, we introduce an augmentation method for positive samples based on dropout. Finally, we propose positive-none-class mixup to further enhance the model's generalization. The overall architecture of P$^{3}$M is shown in Figure \ref{fig1}.

\subsection{Positive-Unlabeled Metric Learning for Document-Level RE}

Document-level RE can be viewed as a multi-label classification task, and there are a large number of entity pairs with no relation. Previous work ~\cite{DBLP:conf/ijcai/ZhouL22,DBLP:conf/emnlp/WangLHZ22} has shown that setting an additional none-class relation can be very helpful for performance. Therefore, in our method, we transform document-level RE with none-class relation into a proxy-based metric learning task, setting an anchor for each positive relation and none-class relation, respectively.

Let $\mathcal{X}$ be an instance space and $\mathcal{Y}=\{-1,+1\}^{K}$ be a label space, where $K$ is the number of pre-defined classes. An instance $\boldsymbol{x} \in \mathcal{X}$ is associated with a subset of labels, identified by a binary vector $\boldsymbol{y} \in \mathcal{Y}=\left(y_{1}, \ldots, y_{K}\right)$, where $y_{i}=+1$ if the $i$-th label is positive for $\boldsymbol{x}$, and $y_{i}=-1$ otherwise. We define each relation embedding as $\boldsymbol{c} \in \mathcal{C}=\left(\boldsymbol{\boldsymbol{c_{0}}}, \boldsymbol{c_{1}}, \ldots, \boldsymbol{c_{K}}\right)$, where $\boldsymbol{c_{0}}$ is the none-class relation embedding, and the rest are predefined relation embeddings. The goal is to learn an embedding $f : \mathcal{X} \rightarrow \mathbb{R}^d$ that brings it closer to its corresponding relation embedding $\boldsymbol{c_{i}}$ and push it further away from the none-class relation embedding $\boldsymbol{c_{0}}$. 

For simplicity, we use the SoftMax$_{norm}$ proposed by \citep{DBLP:conf/iccv/QianSSHTLJ19} as the metric learning loss function, which can be seen as a smoothed version of triplet loss. For a given entity pair and a given relation, the loss function can be expressed as:
\begin{equation}
\begin{aligned}\label{eq1}
&\ell_{SoftMax_{norm}}(f(\boldsymbol{x}), \boldsymbol{c_{i}}, \boldsymbol{c_{0}})= \\ &-\log \frac{\exp (\lambda \boldsymbol{c_{i}}^{\top} f(\boldsymbol{x}))}{\exp (\lambda \boldsymbol{c_{i}}^{\top} f(\boldsymbol{x})) + \exp (\lambda \boldsymbol{c_{0}}^{\top} f(\boldsymbol{x}))},
\end{aligned}
\end{equation}
where $f(\boldsymbol{x})$, $\boldsymbol{c_{i}}$, $\boldsymbol{c_{0}}$ need to be normalized, and $\lambda$ is a scaling factor.

In the inference stage, for any entity pair, the relation $i$ exists if $\boldsymbol{c_{i}}^{\top} f(\boldsymbol{x}) > \boldsymbol{c_{0}}^{\top} f(\boldsymbol{x})$ and vice versa. In the following part we use $\ell$ instead of $\ell_{SoftMax_{norm}}$ as an abbreviation.

For $i$-th class, assume that the data follow an unknown probability distribution with density $p(\boldsymbol{x}, y_{i})$, $p_{\mathrm{P}_{i}}=p(\boldsymbol{x} \mid y_{i}=+1)$ as the positive marginal, $p_{\mathrm{N}_{i}}=p(\boldsymbol{x} \mid y_{i}=-1)$ as the negative marginal, and $p_{i}(\boldsymbol{x})$ as the marginal. In positive-negative metric learning (PNM), the ideal loss to optimize would be:
\begin{equation}
\begin{aligned}\label{eq3}
L_{\mathrm{PNM}}&=\sum_{i=1}^{K}(\pi_{i} \mathbb{E}_{\mathrm{P}_{i}}[\ell(f(\boldsymbol{x}), \boldsymbol{c_{i}}, \boldsymbol{c_{0}})]\\&+(1-\pi_{i})\mathbb{E}_{\mathrm{N}_{i}}[\ell(f(\boldsymbol{x}), \boldsymbol{c_{0}}, \boldsymbol{c_{i}})]),
\end{aligned}
\end{equation}
where $\pi_{i}=p(y_{i}=+1)$ and $(1-\pi_{i})=(1-p(y_{i}=+1))=p(y_{i}=-1)$ is the positive and negative prior of the $i$-th class. $\mathbb{E}_{\mathrm{P}_{i}}[\cdot]=\mathbb{E}_{\boldsymbol{x} \sim p(\boldsymbol{x} \mid y_{i}=+1)}[\cdot]$, $\mathbb{E}_{\mathrm{N}_{i}}[\cdot]=\mathbb{E}_{\boldsymbol{x} \sim p(\boldsymbol{x} \mid y_{i}=-1)}[\cdot]$. 

In positive-unlabeled metric learning (PUM), due to the absence of negative samples, we cannot  estimate $\mathbb{E}_{\mathrm{N}_{i}}[\cdot]$ from the data. Following ~\cite{DBLP:conf/nips/PlessisNS14}, PU learning assumes that unlabeled data can reflect the true overall distribution, that is, $p_{\mathrm{U}_{i}}(\boldsymbol{x})=p_{i}(\boldsymbol{x})$. The expected loss formulation can be defined as:
\begin{equation}
\begin{aligned}\label{eq5}
&L_{\mathrm{PUM}}=\sum_{i=1}^{K}(\pi_{i} \mathbb{E}_{\mathrm{P}_{i}}[\ell(f(\boldsymbol{x}), \boldsymbol{c_{i}}, \boldsymbol{c_{0}})]\\&+\mathbb{E}_{\mathrm{U}_{i}}[\ell(f(\boldsymbol{x}), \boldsymbol{c_{0}}, \boldsymbol{c_{i}})]-\pi_{i} \mathbb{E}_{\mathrm{P}_{i}}[\ell(f(\boldsymbol{x}), \boldsymbol{c_{0}}, \boldsymbol{c_{i}})]),
\end{aligned}
\end{equation}
here $\mathbb{E}_{\mathrm{U}_{i}}[\cdot]=\mathbb{E}_{\boldsymbol{x} \sim p_{i}(\boldsymbol{x})}[\cdot]$ and $\mathbb{E}_{\mathrm{U}_{i}}[\ell(f(\boldsymbol{x}), \boldsymbol{c_{0}}, \boldsymbol{c_{i}})]$ $-\pi_{i} \mathbb{E}_{\mathrm{P}_{i}}[\ell(f(\boldsymbol{x}), \boldsymbol{c_{0}}, \boldsymbol{c_{i}})]$ can alternatively represent $(1-\pi_{i})\mathbb{E}_{\mathrm{N}_{i}}[\ell(f(\boldsymbol{x}), \boldsymbol{c_{0}}, \boldsymbol{c_{i}})]$ because $p_{i}(\boldsymbol{x})=
\pi_{i}p_{\mathrm{P}_{i}}(\boldsymbol{x})+(1-\pi_{i})p_{\mathrm{N}_{i}}(\boldsymbol{x})$.

Since there are already some labeled relations in the document-level RE dataset, this leads to prior shift in the unlabeled data. We also use the method of prior shift in the training data to obtain the final \underline{p}ositive-unlabeled \underline{m}etric learning (PM) expected loss:
\begin{equation}
\begin{aligned}\label{eq6}
L_{\mathrm{PM}}&=\sum_{i=1}^{K}(\pi_{i} \mathbb{E}_{\mathrm{P}_{i}}[ \ell(f(\boldsymbol{x}), \boldsymbol{c_{i}}, \boldsymbol{c_{0}})] \\
&+\frac{1-\pi_{i}}{1-\pi_{u,i}} \mathbb{E}_{\mathrm{U}_{i}}[\ell(f(\boldsymbol{x}), \boldsymbol{c_{0}}, \boldsymbol{c_{i}})] \\
&-\frac{\pi_{u,i}-\pi_{u,i} \pi_{i}}{1-\pi_{u,i}}\mathbb{E}_{\mathrm{P}_{i}}[\ell(f(\boldsymbol{x}), \boldsymbol{c_{0}}, \boldsymbol{c_{i}})]),
\end{aligned}
\end{equation}
where $\pi_{u,i}=p(y_{i}=1 \mid s_{i}=-1)=\frac{\pi_{i}-\pi_{labeled,i}}{1-\pi_{labeled,i}}$, $\pi_{labeled,i}=p(s_{i}=+1)$ and $(1-\pi_{labeled,i})=(1-p(s_{i}=+1))=p(s_{i}=-1)$. $s_{i}=+1$ or $s_{i}=-1$ mean that the $i$-th class is labeled or unlabeled, respectively. For details on prior shift in document-level RE, please refer to ~\cite{DBLP:conf/emnlp/WangLHZ22}.

As a result, by rewriting Eq.\ref{eq6} in the form of data approximation and applying non-negative risk estimation ~\cite{DBLP:conf/nips/KiryoNPS17} to the PM framework to address the overfitting problem caused by the complexity of the model, we can obtain:
\begin{equation}
\begin{aligned}\label{eq7}
&\widehat{L}_{\mathrm{PM}}=\sum_{i=1}^{K}( \frac{1}{n_{\mathrm{P}_{i}}}\pi_{i} \sum_{j=1}^{n_{\mathrm{P}_{i}}}\ell(f(\boldsymbol{x}_{j}^{\mathrm{P}_{i}}), \boldsymbol{c_{i}}, \boldsymbol{c_{0}}) \\&+\mathrm{max}(0, [\frac{1}{n_{\mathrm{U}_{i}}} \frac{1-\pi_{i}}{1-\pi_{u,i}} \sum_{j=1}^{n_{\mathrm{U}_{i}}} \ell(f(\boldsymbol{x}_{j}^{\mathrm{U}_{i}}), \boldsymbol{c_{0}}, \boldsymbol{c_{i}})\\&-\frac{1}{n_{\mathrm{P}_{i}}}\frac{\pi_{u,i}-\pi_{u,i} \pi_{i}}{1-\pi_{u,i}} \sum_{j=1}^{n_{\mathrm{P}_{i}}}\ell(f(\boldsymbol{x}_{j}^{\mathrm{P}_{i}}), \boldsymbol{c_{0}}, \boldsymbol{c_{i}})])),
\end{aligned}
\end{equation}
where $\boldsymbol{x}_{j}^{\mathrm{P}_{i}}$ and $\boldsymbol{x}_{j}^{\mathrm{U}_{i}}$ denote cases that the $j$-th sample of class $i$ is positive or unlabeled. $n_{\mathrm{P}_{i}}$ and $n_{\mathrm{U}_{i}}$ are the number of positive and unlabeled samples of class $i$, respectively. In addition, in order to address the severe class imbalance problem, we multiply the first term in Eq.\ref{eq7} by $\gamma_{i}=(\frac{1-\pi_{i}}{\pi_{i}})^{0.5}$ as the class weight.

\subsection{Positive Augmentation Based on Dropout Noise}

It is important to note that since the labeled sample is only a portion of the true positive sample, meaning the distribution is biased, $p(\boldsymbol{x} \mid y_{i}=1)$ is not equal to $p(\boldsymbol{x} \mid s_{i}=1)$. This means that the first term in Eq.\ref{eq7} is a biased approximation of the first term in Eq.\ref{eq6}. To alleviate this issue, we can use data augmentation to expand the distribution of positive samples. Here, inspired by ~\cite{DBLP:conf/emnlp/GaoYC21}, we use the model's own dropout perturbation to augment the positive samples to get $\boldsymbol{x'}$, and the perturbed entity pair embedding is $f(\boldsymbol{x'})$. However, since we have only augmented the positive samples, the prior $\pi_{u,i}$ in the unlabeled data does not change, and we can obtain the \underline{p}ositive-augmentation \underline{p}ositive-unlabeled \underline{m}etric learning (P$^2$M) objective loss:
\begin{equation}
\begin{aligned}\label{eq8}
L_{\mathrm{P^{2}M}}&=\sum_{i=1}^{K}(\pi_{i} \mathbb{E}_{\mathrm{P}_{new,i}}[ \ell(f(\boldsymbol{x}), \boldsymbol{c_{i}}, \boldsymbol{c_{0}})] \\
&+\frac{1-\pi_{i}}{1-\pi_{u,i}} \mathbb{E}_{\mathrm{U}_{i}}[\ell(f(\boldsymbol{x}), \boldsymbol{c_{0}}, \boldsymbol{c_{i}})] \\
&-\frac{\pi_{u,i}-\pi_{u,i} \pi_{i}}{1-\pi_{u,i}}\mathbb{E}_{\mathrm{P}_{new,i}}[\ell(f(\boldsymbol{x}), \boldsymbol{c_{0}}, \boldsymbol{c_{i}})]),
\end{aligned}
\end{equation}
here $\mathbb{E}_{\mathrm{P}_{new,i}}[\cdot]=\mathbb{E}_{\boldsymbol{x},\boldsymbol{x'} \sim p_{i}(\boldsymbol{x},\boldsymbol{x'} \mid y_{i}=+1)}[\cdot]$. It can be written in the non-negative form of data approximation as:
\begin{equation}
\begin{aligned}\label{eq9}
&\widehat{L}_{\mathrm{P^{2}M}}=\sum_{i=1}^{K}( \frac{1}{2n_{\mathrm{P}_{i}}}\pi_{i} (\sum_{j=1}^{n_{\mathrm{P}_{i}}}\ell(f(\boldsymbol{x}_{j}^{\mathrm{P}_{i}}), \boldsymbol{c_{i}}, \boldsymbol{c_{0}}) \\&+ \sum_{j=1}^{n_{\mathrm{P}_{i}}}\ell(f(\boldsymbol{x'}_{j}^{\mathrm{P}_{i}}), \boldsymbol{c_{i}}, \boldsymbol{c_{0}})) \\&+\mathrm{max}(0, [\frac{1}{n_{\mathrm{U}_{i}}} \frac{1-\pi_{i}}{1-\pi_{u,i}} \sum_{j=1}^{n_{\mathrm{U}_{i}}} \ell(f(\boldsymbol{x}_{j}^{\mathrm{U}_{i}}), \boldsymbol{c_{0}}, \boldsymbol{c_{i}})\\&-\frac{1}{2n_{\mathrm{P}_{i}}}\frac{\pi_{u,i}-\pi_{u,i} \pi_{i}}{1-\pi_{u,i}} (\sum_{j=1}^{n_{\mathrm{P}_{i}}}\ell(f(\boldsymbol{x}_{j}^{\mathrm{P}_{i}}), \boldsymbol{c_{0}}, \boldsymbol{c_{i}}) \\&+ \sum_{j=1}^{n_{\mathrm{P}_{i}}}\ell(f(\boldsymbol{x'}_{j}^{\mathrm{P}_{i}}), \boldsymbol{c_{0}}, \boldsymbol{c_{i}}))])).
\end{aligned}
\end{equation}

We will compare the difference between augmenting positive samples and augmenting all samples in the main results subsection of the experiments.

\subsection{Positive-None-Class Mixup}

In order to further enhance the generalization of the model, our goal is to obtain more diverse entity pair embeddings. To achieve this, we can interpolate between positive and negative samples of each class to obtain mixed entity pair representations:
\begin{equation}
\begin{aligned}\label{eq10}
f_{mix(ori)}(\boldsymbol{x})=\mu f(\boldsymbol{x})+(1-\mu) f(\boldsymbol{x}_{-}),
\end{aligned}
\end{equation}
here $\boldsymbol{x} \sim p(\boldsymbol{x} \mid y_{i}=1)$ and $\boldsymbol{x}_{-} \sim p(\boldsymbol{x} \mid y_{i}=-1)$ are the positive and negative samples of class $i$, respectively. $\mu$ is sampled from
a $\mathrm{Beta}(\alpha, \alpha)$ distribution ($\mu \in [0, 1]$ and $\alpha > 0$). However, in PU learning, we cannot obtain true negative samples, which means that there are some positive entity pairs in the unlabeled samples, resulting in bias when interpolating with unlabeled entity pairs. Thanks to the metric learning framework, which places the relation embedding and the entity pair embedding in the same feature space, we can use the none-class relation embedding $\boldsymbol{c_{0}}$ to stand in for pseudo-negative entity pairs. Therefore, Eq.\ref{eq10} can be rewritten as:
\begin{equation}
\begin{aligned}\label{eq11}
f_{mix}(\boldsymbol{x})=\mu f(\boldsymbol{x})+(1-\mu) \boldsymbol{c_{0}}.
\end{aligned}
\end{equation}

We will compare the difference between using mixup with none-class relation embedding and the original method in the main results subsection of the experiments.
According to this formulation, the mixup loss function can be reformulated as:
\begin{equation}
\begin{aligned}\label{eq12}
L_{p-mix} &= \sum_{i=1}^{K} (\mu \mathbb{E}_{\mathrm{P}_{new,i}}[ \ell(f_{mix}(\boldsymbol{x}), \boldsymbol{c_{i}}, \boldsymbol{c_{0}})] \\&+ (1-\mu) \mathbb{E}_{\mathrm{P}_{new,i}}[ \ell(f_{mix}(\boldsymbol{x}), \boldsymbol{c_{0}}, \boldsymbol{c_{i}})]),
\end{aligned}
\end{equation}
here we perform mixup on both the original positive samples and the augmented positive samples. Finally, we rewrite this equation in the form of data approximation:
\begin{equation}
\begin{aligned}\label{eq13}
\widehat{L}_{p-mix} &= \sum_{i=1}^{K} (\frac{\mu}{2n_{\mathrm{P}_{i}}} (\sum_{j=1}^{n_{\mathrm{P}_{i}}}\ell(f_{mix}(\boldsymbol{x}_{j}^{\mathrm{P}_{i}}), \boldsymbol{c_{i}}, \boldsymbol{c_{0}}) \\&+ \sum_{j=1}^{n_{\mathrm{P}_{i}}} \ell(f_{mix}(\boldsymbol{x'}_{j}^{\mathrm{P}_{i}}), \boldsymbol{c_{i}}, \boldsymbol{c_{0}})) \\&+ \frac{(1-\mu)}{2n_{\mathrm{P}_{i}}} (\sum_{j=1}^{n_{\mathrm{P}_{i}}}\ell(f_{mix}(\boldsymbol{x}_{j}^{\mathrm{P}_{i}}), \boldsymbol{c_{0}}, \boldsymbol{c_{i}}) \\&+ \sum_{j=1}^{n_{\mathrm{P}_{i}}} \ell(f_{mix}(\boldsymbol{x'}_{j}^{\mathrm{P}_{i}}), \boldsymbol{c_{0}}, \boldsymbol{c_{i}}))).
\end{aligned}
\end{equation}

Therefore, the final loss of our \underline{p}ositive-augmentation and \underline{p}ositive-mixup \underline{p}ositive-unlabeled \underline{m}etric learning (P$^{3}$M) framework is:
\begin{equation}
\begin{aligned}\label{eq14}
\widehat{L}_{P^{3}M} = \widehat{L}_{P^{2}M} + \nu \widehat{L}_{p-mix},
\end{aligned}
\end{equation}
where $\nu$ is a hyperparameter that controls the strength of positive-mixup.

\begin{table*}[!ht]
\centering
\begin{tabular}{lcccccc}
\hline \multirow{2}{*}{ Dataset } & \multicolumn{1}{c}{DocRED} & \multicolumn{1}{c}{DocRED\_ext} & \multicolumn{2}{c}{Re-DocRED} & \multicolumn{2}{c}{ChemDisGene} \\
 & train & train & train & test & train & test \\
\hline 
\# docs & 3,053 & 3,053 & 3,053 & 500 & 76,942 & 523 \\
\# rels & 96 & 96 & \multicolumn{2}{c}{96} & \multicolumn{2}{c}{14} \\
Avg \# ents & $19.5$ & $19.5$ & $19.4$ & $19.6$ & $7.5$ & $10.0$ \\
Avg \# rels & $12.5$ & $5.4$ & $28.1$ & $34.9$ & $2.1$ & $7.2$ \\
\hline
\end{tabular}
\caption{Statistics of document-level RE datasets.}
\label{tab1}
\end{table*}

\begin{table*}[!ht]
\centering
\begin{tabular}{lcccc|cccc}
\toprule
\multirow{2}{*}{\textbf{Model}} & \multicolumn{4}{c}{\textbf{DocRED}} & \multicolumn{4}{c}{\textbf{DocRED\_ext}} \\
\cmidrule(l){2-9} & Ign F1 & F1 & P & R & Ign F1 & F1 & P & R \\ \midrule
BiLSTM$^{\dagger}$ & $32.57$ & $32.86$ & $77.04$ & $20.89$ & $-$ & $-$ & $-$ & $-$ \\
GAIN+BERT$_{Base}^{\dagger}$ & $45.57$ & $45.82$ & $88.11$ & $30.98$ & $-$ & $-$ & $-$ & $-$ \\
DocuNET+RoBERTa$_{Large}^{\dagger}$ & $45.88$ & $45.99$ & $94.16$ & $30.42$ & $-$ & $-$ & $-$ & $-$ \\
\midrule
ATLOP+BERT$_{Base}^{\dagger}$ & $43.12$ & $43.25$ & $\mathbf{92.49}$ & $28.23$ & $16.99$ & $17.01$ & $\mathbf{93.17}$ & $9.36$ \\
SSR-PU+ATLOP+BERT$_{Base}^{\dagger}$ & $55.21$ & $56.14$ & $70.42$ & $46.67$ & $46.47$ & $47.24$ & $59.52$ & $39.18$ \\
PM+ATLOP+BERT$_{Base}$ & $57.97$ & $59.34$ & $60.76$ & $58.01$ & $53.84$ & $54.85$ & $54.91$ & $\mathbf{54.81}$ \\
P$^{2}$M(all)+ATLOP+BERT$_{Base}$ & $58.27$ & $59.54$ & $63.31$ & $56.19$ & $53.77$ & $54.71$ & $56.81$ & $52.78$ \\
P$^{2}$M+ATLOP+BERT$_{Base}$ & $58.85$ & $60.08$ & $64.30$ & $56.40$ & $54.64$ & $55.57$ & $58.10$ & $53.26$ \\
P$^{3}$M(ori)+ATLOP+BERT$_{Base}$ & $59.48$ & $60.79$ & $62.53$ & $\mathbf{59.14}$ & $55.91$ & $56.82$ & $59.13$ & $54.70$ \\
P$^{3}$M+ATLOP+BERT$_{Base}$ & $\mathbf{59.81}$ & $\mathbf{61.03}$ & $64.57$ & $57.87$ & $\mathbf{56.17}$ & $\mathbf{57.02}$ & $61.12$ & $53.44$ \\
\midrule
ATLOP+RoBERTa$_{Large}^{\dagger}$ & $45.09$ & $45.19$ & $\mathbf{94.75}$ & $29.67$ & $17.29$ & $17.31$ & $\mathbf{94.85}$ & $9.52$ \\
SSR-PU+ATLOP+RoBERTa$_{Large}^{\dagger}$ & $58.68$ & $59.50$ & $74.21$ & $49.67$ & $48.98$ & $49.74$ & $61.57$ & $41.75$ \\
PM+ATLOP+RoBERTa$_{Large}$ & $60.72$ & $62.13$ & $61.15$ & $63.16$ & $56.67$ & $57.72$ & $54.60$ & $\mathbf{61.27}$ \\
P$^{2}$M(all)+ATLOP+RoBERTa$_{Large}$ & $61.17$ & $62.46$ & $64.19$ & $60.83$ & $56.49$ & $57.46$ & $57.42$ & $57.51$ \\
P$^{2}$M+ATLOP+RoBERTa$_{Large}$ & $61.55$ & $62.82$ & $65.19$ & $60.62$ & $57.27$ & $58.21$ & $58.91$ & $57.55$ \\
P$^{3}$M(ori)+ATLOP+RoBERTa$_{Large}$ & $62.64$ & $63.96$ & $64.15$ & $\mathbf{63.80}$ & $58.48$ & $59.42$ & $59.53$ & $59.33$ \\
P$^{3}$M+ATLOP+RoBERTa$_{Large}$ & $\mathbf{63.16}$ & $\mathbf{64.34}$ & $67.43$ & $61.52$ & $\mathbf{59.02}$ & $\mathbf{59.86}$ & $63.04$ & $57.01$ \\
\bottomrule
\end{tabular}
\caption{
Results on Re-DocRED revised test set. Results with $\dagger$ are reported from ~\cite{DBLP:conf/emnlp/WangLHZ22}.
}
\label{tab2}
\end{table*}

\section{Experiments}
In this section, we evaluate the performance of P$^{3}$M in various incompletely labeled document-level RE datasets and settings as well as in the fully labeled scenario. We also analyze the effectiveness of different components of the method.

\subsection{Experimental Setups}

\subsubsection{Datasets.}
\emph{DocRED} ~\cite{DBLP:conf/acl/YaoYLHLLLHZS19} is a large-scale document-level RE dataset constructed from Wikipedia, containing 96 predefined relations. However, the original dataset contains a large amount of incomplete labeling phenomena, ~\cite{DBLP:conf/emnlp/Tan0BNA22} proposed a high-quality revised version Re-DocRED. In our experiments, we use the incompletely labeled DocRED original training set and the fully labeled Re-DocRED test set. In order to further analyze the performance of the method in incompletely labeled scenarios, we also use the extreme incompletely labeled training set DocRED\_ext constructed by ~\cite{DBLP:conf/emnlp/WangLHZ22} for experiments. \enspace \emph{ChemDisGene} ~\cite{DBLP:conf/lrec/ZhangMTM22} is a multi-label document-level RE dataset in the biomedical field. We use the incompletely labeled training set constructed by distantly supervised of CTD database ~\cite{DBLP:journals/nar/DavisGJSWWM21} and the fully labeled \emph{All relationships} test set constructed by additional annotation by domain experts for experiments. \enspace The datasets are in English and used for their intended purpose. The detailed statistics of the datasets are shown in Table \ref{tab1}. The average number of relations in the incompletely labeled training sets, especially the extremely incompletely labeled sets, is far less than that in the test sets, indicating the large number of false negatives in the training sets.

\subsubsection{Implementation Details.}
In our experiments, we use ATLOP ~\cite{DBLP:conf/aaai/Zhou0M021} as the encoding model for relation representation learning. Further, we apply cased $\mathrm{BERT}_{Base}$ ~\cite{DBLP:conf/naacl/DevlinCLT19} and $\mathrm{RoBERTa}_{Large}$ ~\cite{DBLP:journals/corr/abs-1907-11692} for DocRED and $\mathrm{PubmedBert}$ ~\cite{DBLP:journals/health/GuTCLULNGP22} for ChemDisGene. We use Huggingface's Transformers ~\cite{DBLP:conf/emnlp/WolfDSCDMCRLFDS20} to implement all the models and AdamW ~\cite{DBLP:conf/iclr/LoshchilovH19} as the optimizer, and apply a linear warmup ~\cite{DBLP:journals/corr/GoyalDGNWKTJH17} at the first 6\% steps followed by a linear decay to 0. For DocRED, we set the learning rates to 3e-5. For ChemDisGene, the learning rate is set to 2e-5. The batch size (number of documents per batch) is set to 4 and 8 for two datasets, respectively. In our experiments, we fixed $\lambda=10$, $\alpha=1.0$, $\nu=0.05$, and the dropout rate to 0.2. In order to make a fair comparison, we use the same prior estimation as in ~\cite{DBLP:conf/emnlp/WangLHZ22}, setting $\pi_{i}=3\pi_{labeled,i}$ and for extreme incomplete labeling, setting $\pi_{i}=12\pi_{labeled,i}$. The training stopping criteria are set as follows: 10 epochs for both two dataset. We do not use any fully labeled validation or test sets in any stage of the training process and report the results of the final model by running five times with different random seeds (62, 63, 64, 65, 66). All experiments are conducted with 1 Tesla A100 or 1 Tesla V100 GPU.

\subsubsection{Baseline.}
For DocRED, we use fully supervised models BiLSTM ~\cite{DBLP:conf/acl/YaoYLHLLLHZS19}, GAIN ~\cite{DBLP:conf/emnlp/ZengXCL20}, DocuNET ~\cite{DBLP:conf/ijcai/ZhangCXDTCHSC21}, and ATLOP ~\cite{DBLP:conf/aaai/Zhou0M021}, as well as the positive-unlabeled learning method SSR-PU ~\cite{DBLP:conf/emnlp/WangLHZ22} as the baseline models. For ChemDisGene, we use BRAN ~\cite{DBLP:conf/naacl/VergaSM18}, PubmedBert ~\cite{DBLP:journals/health/GuTCLULNGP22}, PubmedBert+BRAN ~\cite{DBLP:conf/lrec/ZhangMTM22}, ATLOP, and SSR-PU as the baselines.

\subsubsection{Evaluation Metric.}
We use micro F1 (F1), micro ignore F1 (Ign F1), precision (P), and recall (R) to evaluate the overall performance of models on DocRED. Ign F1 denotes the F1 score excluding the relations shared by the training and test set. We use micro F1 (F1), precision (P), and recall (R) to evaluate the models on ChemDisGene.

\begin{table}[!ht]
\centering
\begin{tabular}{lccc}
\toprule
\textbf{Model} & F1 & P & R \\
\midrule BRAN$^{\ddagger}$ & $32.5$ & $41.8$ & $26.6$ \\
PubmedBert$^{\ddagger}$ & $42.1$ & $64.3$ & $31.3$ \\
BRAN+PubmedBert$^{\ddagger}$ & $43.8$ & $70.9$ & $31.6$ \\
\midrule
ATLOP+PubmedBert$^{\dagger}$ & $42.73$ & $\mathbf{76.17} $ & $29.70$ \\
SSR-PU+PubmedBert$^{\dagger}$ & $48.56$ & $54.27$ & $43.93$ \\
PM+PubmedBert & $52.02$ & $58.26$ & $47.00$ \\
P$^{2}$M(all)+PubmedBert & $51.29$ & $57.54$ & $46.27$ \\
P$^{2}$M+PubmedBert & $52.19$ & $59.02$ & $46.78$ \\
P$^{3}$M(ori)+PubmedBert & $53.58$ & $59.44$ & $\mathbf{48.78}$ \\
P$^{3}$M+PubmedBert & $\mathbf{53.62}$ & $60.20$ & $48.34$ \\
\bottomrule
\end{tabular}
\caption{
Results on ChemDisGene \emph{All relationships} test set. Results with $\dagger$ are reported from ~\cite{DBLP:conf/emnlp/WangLHZ22}. Results with $\ddagger$ are reported from ~\cite{DBLP:conf/lrec/ZhangMTM22}.
}
\label{tab3}
\end{table}

\begin{table}
\centering
\begin{tabular}{lcc}
\toprule \textbf{Model} & Ign F1 & F1 \\
\midrule ATLOP+BERT$_{Base}^{\dagger}$ & $72.70$ & $73.47$ \\
SSR-PU+BERT$_{Base}^{\dagger}$ & $72.91$ & $74.33$ \\
P$^{3}$M+BERT$_{Base}$ & $\mathbf{74.10}$ & $\mathbf{75.60}$ \\
\midrule ATLOP+RoBERTa$_{Large}^{\dagger}$ & $76.92$ & $77.58$ \\
DocuNET+RoBERTa$_{Large}^{\ddagger}$ & $77.26$ & $77.87$ \\
KD-DocRE+RoBERTa$_{Large}^{\ddagger}$ & $77.60$ & $78.28$ \\
SSR-PU+RoBERTa$_{Large}^{\dagger}$ & $77.67$ & $78.86$ \\
P$^{3}$M+RoBERTa$_{Large}$ & $\mathbf{78.82}$ & $\mathbf{80.02}$ \\
\bottomrule
\end{tabular}
\caption{Results on Re-DocRED revised test set under the fully supervised setting. Results with $\dagger$ are reported from ~\cite{DBLP:conf/emnlp/WangLHZ22}. Results with $\ddagger$ are reported from ~\cite{DBLP:conf/emnlp/Tan0BNA22}.
}
\label{tab4}
\end{table}

\subsection{Main Results}\label{sec4.2}
In this subsection, we compare the results between PM, P$^{2}$M(all), P$^{2}$M, P$^{3}$M(ori), and P$^{3}$M. P$^{2}$M(all) refers to augmenting all samples, while P$^{3}$M(ori) refers to the original mixup method that uses unlabeled samples for mixing.

\subsubsection{Results on DocRED.}
As shown in Table \ref{tab2}, traditional supervised learning methods such as BiLSTM, GAIN, DocuNET, and ATLOP have a dramatic decline in performance, especially in recall, in the incompletely labeled scenario. The SSR-PU method, which uses PU learning, effectively alleviates this problem and achieves a huge improvement on the basis of the ATLOP encoder model. Our framework P$^{3}$M, on the other hand, improves the F1 score by 4.89 and 4.84, respectively, for the $\mathrm{BERT}_{Base}$ and $\mathrm{RoBERTa}_{Large}$ settings, compared to SSR-PU in the incompletely labeled scenario. And when using extremely incompletely labeled training sets, the two settings respectively improve the F1 score by 9.78 and 10.12. The outstanding improvement shows the effectiveness of our proposed framework.

In the DocRED experiment, P$^{3}$M shows a slight precision drop compared to SSR-PU, possibly due to expanded positive sample distribution causing errors in ambiguous case classification. However, recall increases of 11.20 and 11.85 under the $\mathrm{BERT}_{Base}$ and $\mathrm{RoBERTa}_{Large}$ settings justify this trade-off. This classification error, likely stemming from the base model's limitations, can be mitigated by enhancing the base model. In DocRED\_ext, our method not only improves recall by 14.26 and 15.26 over SSR-PU under the same settings but also raises precision by 1.60 and 1.47, respectively, highlighting its value in label-scarce scenarios.

We compare different variations of our method. P$^{2}$M(all) and P$^{2}$M use dropout to augment samples. P$^{2}$M(all) augments all samples, and there is a slight improvement when using the DocRED training set compared to the basic PM framework. However, in extreme scenarios of incomplete labeling, performance deteriorates. We believe this is caused by the fact that since the data itself is positive and unlabeled, augmenting all samples instead introduces some additional noise. P$^{2}$M, which only augments positive samples, does not have this problem. The increase in the distribution of positive samples further improves the performance of the model and to some extent relieves the distribution bias caused by incompletely labeled positive samples. The regular P$^{3}$M(ori) method has a considerable improvement over P$^{2}$M because in document-level RE, the number of negative samples is far greater than that of positive samples. Therefore, direct sampling of unlabeled samples will only introduce a small number of false negatives, but there will still be bias. P$^{3}$M has more performance improvement compared to P$^{3}$M(ori), which shows that using none-class relation embedding as pseudo-negative samples effectively mitigates the bias of directly using unlabeled samples for mixup.

\subsubsection{Results on ChemDisGene.}
As shown in Table \ref{tab3}, similar to the results on DocRED, the performance of supervised learning methods has a large decline, while SSR-PU has a large improvement compared to them. Our proposed P$^{3}$M improved by 5.06 F1 score compared to SSR-PU and achieved a new best result. Notably, the \emph{All relationships} test set of ChemDisGene is sourced from another corpus DrugProt ~\cite{miranda2021overview} and is additionally annotated by human experts, making the test set have a larger deviation from the training set. However, our proposed framework has better robustness under this deviation.

For different variations of the method, P$^2$M(all) has an obvious performance decrease compared to PM, which may be caused by the larger deviation between the training set and the test set, and this deviation is further amplified by the augmentation of all samples. P$^2$M shows further improvement compared to PM, indicating the help of dropout augmentation in improving the diversity of positive sample distribution. P$^3$M(ori) and P$^3$M, which added positive-mixup method, both have larger improvements, verifying the help of mixup in improving the generalization of positive-unlabeled metric learning framework. Due to the scarcity of positive samples, sampling $n_{\mathrm{P}_{i}}$ as negative samples from the unlabeled samples only causes a small bias, making P$^3$M(ori) still able to achieve a good result and the performance gap between P$^3$M and P$^3$M(ori) is smaller.

\begin{figure}
\centering
\includegraphics[width=0.47\textwidth]{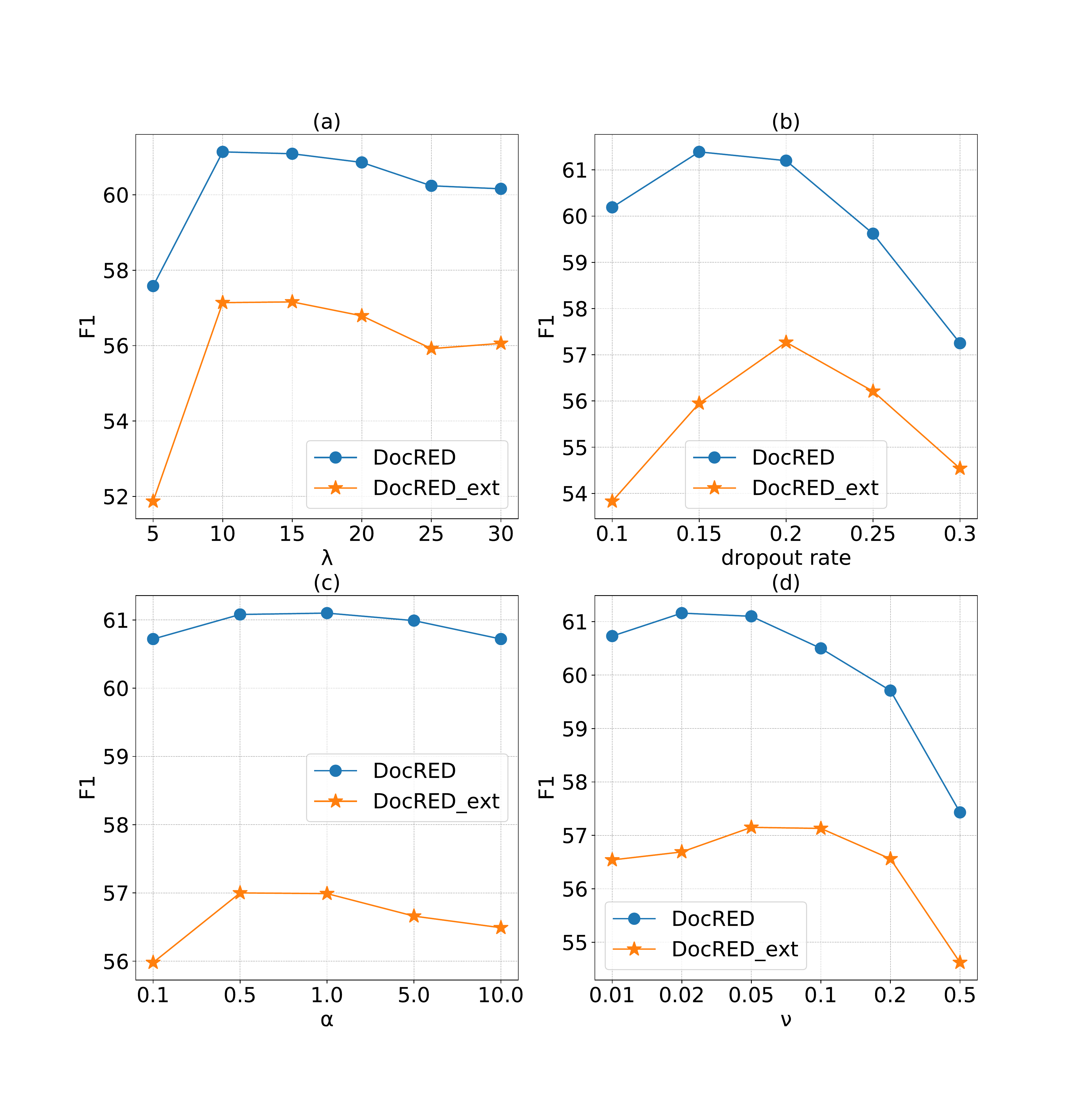}
\caption{Effect of hyperparameters on DocRED}
\label{fig2}
\end{figure}

\begin{table}
\centering
\begin{tabular}{lcccc}
\toprule \textbf{Model} & Ign F1 & F1 & P & R \\
\midrule
P$^{3}$M$_{\pi_{i}=\pi_{labeled,i}}$& $59.84$ & $60.77$ & $70.99$ & $53.12$ \\
P$^{3}$M$_{\pi_{i}=2\pi_{labeled,i}}$ & $60.00$ & $61.06$ & $68.06$ & $55.38$ \\
P$^{3}$M$_{\pi_{i}=3\pi_{labeled,i}}$ & $59.92$ & $61.13$ & $65.01$ & $57.69$ \\
P$^{3}$M$_{\pi_{i}=4\pi_{labeled,i}}$ & $59.06$ & $60.43$ & $61.08$ & $59.78$ \\
P$^{3}$M$_{\pi_{i}=5\pi_{labeled,i}}$ & $57.98$ & $59.48$ & $57.75$ & $61.31$ \\
\bottomrule
\end{tabular}
\caption{
Results on Re-DocRED revised test set under the $\mathrm{BERT}_{Base}$ setting with different $\pi_{i}$ estimation.
}
\label{tab5}
\end{table}

\subsection{Additional Analysis}

\subsubsection{Fully Supervised Setting.}
We conduct experiments on Re-DocRED ~\cite{DBLP:conf/emnlp/Tan0BNA22} under a fully supervised setting. In the experiments, we set $\pi_{i}=\pi_{labeled,i}$, $\nu=0.01$, with a dropout rate of 0.1, and other hyperparameters remained unchanged. As shown in Table \ref{tab4}, we compared our framework with the existing state-of-the-art methods ATLOP ~\cite{DBLP:conf/aaai/Zhou0M021}, DocuNET ~\cite{DBLP:conf/ijcai/ZhangCXDTCHSC21}, KD-DocRE ~\cite{DBLP:conf/acl/TanHBN22}, and SSR-PU \cite{DBLP:conf/emnlp/WangLHZ22}. Our framework achieves the best results as well.

\subsubsection{Effect of Hyperparameters.}
Figure \ref{fig2} shows the effect of hyperparameters on the model under the DocRED and DocRED\_ext incomplete labeling settings. (a) shows the effect of the scaling factor $\lambda$, with similar trends under both settings, and $\lambda=10$ being the best choice. (b) shows the effect of the dropout rate, indicating that the greater the degree of incompleteness in labeling, the greater the dropout rate needed to enhance diversity, but too large a dropout rate will also introduce more noise. (c) shows the effect of $\alpha$, indicating that the model is not sensitive to the choice of $\alpha$, and $\alpha=1.0$ can be seen as a uniform mixup interpolation between distributions. (d) shows the effect of $\nu$, with similar trends under both settings, and more severe incomplete labeling requires slightly larger mixup strength.

\subsubsection{Effect of Prior Estimation.}
Table \ref{tab5} shows the effect of different prior estimates on the model. It can be seen that our framework is not sensitive to errors in prior estimates, especially in cases where the prior estimate is too small. Even when $\pi_{i}=\pi_{labeled,i}$, the model still performs well, demonstrating the robustness of our method under errors in prior estimates, which is very helpful for real-world applications.

\section{Related Work}

\subsubsection{Document-Level Relation Extraction.}
Previously, effective methods for document-level relation extraction (RE) have mainly been graph-based models and transformer-based models. Graph-based models ~\cite{DBLP:conf/acl/NanGSL20,DBLP:conf/coling/LiYSXXZ20,DBLP:conf/emnlp/ZengXCL20, DBLP:conf/acl/ZengWC21, DBLP:conf/aaai/XuCZ21} use graph neural networks to gather entity information for relational inference, while transformer-based methods ~\cite{DBLP:conf/aaai/Zhou0M021,DBLP:conf/aaai/XuWLZM21,DBLP:conf/ijcai/ZhangCXDTCHSC21,DBLP:conf/acl/TanHBN22} capture long-range dependencies implicitly. Recently, it has been found that there are a large number of false negatives in document-level RE datasets, i.e. incomplete labels ~\cite{DBLP:conf/acl/HuangH0ZF022, DBLP:conf/emnlp/Tan0BNA22}. ~\cite{DBLP:conf/emnlp/WangLHZ22} proposed using positive-unlabeled learning to address this problem.

\subsubsection{Positive-Unlabeled Learning.}
Positive-unlabeled (PU) learning ~\cite{DBLP:conf/kdd/ElkanN08,DBLP:conf/nips/PlessisNS14,DBLP:conf/icml/PlessisNS15,DBLP:conf/nips/KiryoNPS17,DBLP:conf/nips/GargWSBL21}, as a emerging weakly supervised learning paradigm, aims to learn classifiers from positive and unlabeled data, and has gained continuous attention from researchers. PU learning has been widely applied in various tasks, such as text classification ~\cite{DBLP:conf/ijcai/LiL03}, sentence embedding ~\cite{DBLP:conf/emnlp/CaoLERMH21}, named entity recognition ~\cite{DBLP:conf/acl/PengXZFH19, DBLP:conf/acl/ZhouLL22}, knowledge graph completion ~\cite{DBLP:conf/ijcai/TangPZZZH022}, and sentence-level RE ~\cite{DBLP:conf/aaai/HeC0ZWZ20} in the NLP field. ~\cite{DBLP:conf/nips/ChuangRL0J20} used PU learning to address the issue of negative samples potentially carrying the same label in contrastive learning.

\subsubsection{Deep Metric Learning.}
Our work is inspired by metric learning and mainly falls into two categories: pair-based losses and proxy-based losses. Pair-based methods ~\cite{DBLP:conf/cvpr/HadsellCL06,DBLP:conf/cvpr/SchroffKP15,DBLP:conf/nips/Sohn16,DBLP:conf/cvpr/WangHHDS19} focus on the relationships between individual samples, and contrastive learning can be considered a subset of this approach. Proxy-based methods like Proxy-NCA ~\cite{DBLP:conf/iccv/Movshovitz-Attias17} and NormFace ~\cite{DBLP:conf/mm/WangXCY17} consider the relationships between proxies and samples, and ~\cite{DBLP:conf/iccv/QianSSHTLJ19} unified the relationship between SoftMax loss and triplet loss, and proposed a new SoftTriplet loss. Proxy-based methods are a type of approach that focuses on improving generalization while keeping training complexity low, although they may not fully utilize the relationships between individual samples.

\subsubsection{Data Augmentation.}
Data augmentation is a key factor in deep learning performance and is widely used in many fields ~\cite{DBLP:journals/jbd/ShortenK19,DBLP:conf/naacl/HedderichLASK21}. ~\cite{DBLP:conf/emnlp/WeiZ19,ma2019nlpaug} proposed to augment words by randomly inserting and replacing them, while ~\cite{DBLP:conf/acl/LeeKLH20} augmented the word embeddings directly. ~\cite{DBLP:conf/emnlp/GaoYC21} used simple dropout to augment sentence embeddings for unsupervised contrastive learning. Mixup ~\cite{DBLP:conf/iclr/ZhangCDL18,DBLP:conf/icml/VermaLBNMLB19} can be considered as another common data augmentation method, where interpolation is used to improve the generalization performance of the model between two samples. It is increasingly used and researched in the NLP ~\cite{DBLP:conf/acl/ChenYY20,DBLP:conf/acl/YinWQX21,DBLP:conf/naacl/WuX0ZMXZ22} and the PU learning ~\cite{DBLP:conf/nips/ChenLWZW20,DBLP:journals/corr/abs-2004-09388,DBLP:conf/iclr/LiLFO22,DBLP:conf/cvpr/ZhaoXJWH22} fields. ~\cite{DBLP:conf/acl/JeongBCHP22} proposed a document augmentation dense retrieval framework that uses both methods.

\section{Conclusion and Future Work}

To address document-level RE with incomplete labeling, we propose a positive-unlabeled metric learning framework P$^{3}$M. First, we combine positive-unlabeled learning with metric learning to learn better representations. Then, we use dropout augmentation to expand the distribution of labeled positive samples. Finally, we use none-class relation embedding as pseudo-negative samples and propose a positive-none-class mixup method to further improve the model's generalization performance. Experiments demonstrate that our method achieve state-of-the-art results in both incomplete and complete labeling scenarios, as well as robustness to prior estimation bias. In the future, we will explore various metric learning losses and data augmentation methods.

\section{Acknowledgments}

This work is funded by National Natural Science Foundation of China (under project No. 62377013).The computation is supported by the ECNU Multifunctional Platform for Innovation (001).

\bibliography{aaai24}

\end{document}